\pdfoutput=1

\documentclass[11pt]{article}

\usepackage{acl}

\usepackage{times}
\usepackage{latexsym}
\usepackage{amssymb}
\usepackage{amsmath}
\usepackage{graphicx}
\usepackage{booktabs}
\usepackage[T1]{fontenc}

\usepackage[utf8]{inputenc}

\usepackage{microtype}

\usepackage{inconsolata}

\newcommand\blfootnote[1]{%
  \begingroup
  \renewcommand\thefootnote{}\footnote{#1}%
  \addtocounter{footnote}{-1}%
  \endgroup
}

\title{Empirical study of pretrained multilingual language models \\ for zero-shot cross-lingual knowledge transfer in generation}

\author{Nadezhda Chirkova \\
  Naver Labs Europe  \\
   Grenoble, France  \\
  \texttt{nadia.chirkova} \\
  \texttt{@naverlabs.com} \\\And
  Sheng Liang\thanks{~~Work done while Sheng Liang was an intern at Naver Labs Europe.} \\
  Center for Information and \\ 
  Language Processing (CIS) \\
  LMU Munich, Germany \\
  \texttt{shengliang@cis.lmu.de} \\\And
  Vassilina Nikoulina \\
  Naver Labs Europe  \\
   Grenoble, France  \\
  \texttt{vassilina.nikoulina} \\
  \texttt{@naverlabs.com} \\}

\begin{document}
\maketitle
\begin{abstract}
Zero-shot cross-lingual knowledge transfer enables the multilingual pretrained language model (mPLM), finetuned on a  task in one language, make predictions for this task in other languages. While being broadly studied for natural language understanding tasks, the described setting is understudied for generation. Previous works notice a frequent problem of generation in a wrong language and propose approaches to address it, usually using mT5 as a backbone model. In this work, we test alternative mPLMs, such as mBART and NLLB-200, considering full finetuning and parameter-efficient finetuning with adapters. We find that mBART with adapters performs similarly to mT5 of the same size, and NLLB-200 can be competitive in some cases. We also underline the importance of tuning learning rate used for finetuning, which helps to alleviate the problem of generation in the wrong language.
\end{abstract}

\section{Introduction}
\blfootnote{Author contributions: Nadezhda Chirkova run experiments for the preprint and wrote text; Sheng Liang run preliminary experiments; Vassilina Nikoulina proposed and scientifically supervised the project.}
Multilingual pretrained language models (mPLMs) such as mBERT~\cite{mbert}, mBART~\cite{mbartpt}, and mT5~\cite{mt5} provide high-quality representations for texts in various languages and serve as a a universal backbone for finetuning on language-specific task data. The latter, however, is not always available for a language of interest, providing motivation for studying \textit{zero-shot cross-lingual} capabilities of mPLMs. In this setting, the model is finetuned on the task data in one \textit{source} language, usually English, and then applied in a zero-shot manner to make predictions in another \textit{target} language, seen only at the pretraining stage. 

While the described setting was broadly studied for natural language understanding tasks~\cite{mt5,commoncrawl,artetxe-etal-2020-cross,pires-etal-2019-multilingual,wu-dredze-2019-beto,pfeiffer-etal-2020-mad}, work on zero-shot cross-lingual \textit{generation} is more limited~\cite{vu-etal-2022-overcoming,mmt5,zmbart,li-murray-2023-zero}. Two main problems arising in this scenario are usually highlighted:
producing incoherent or irrelevant answers, and generating text in a wrong language. A series of potential solutions for the latter problem were proposed, such as freezing parts of the weights during finetuning, mixing-in unsupervised target language data together with supervised source language data, or using language-specific modules in an architecture.

One missing piece in the literature is that most of the works consider only one base model, mT5, ignoring other existing mPLMs. Various models differ in architectural details, pretraining procedure including pretraining task and data, and use of language codes concatenated to input and output sequences. 
For example, the use of target language code in decoding, such as in mBART, may help guiding generation in a correct language, or the use of translation pretraining task, such as in NLLB-200~\cite{nllb}, may provide well transferable representations.

In this work, we conduct an empirical study of alternatives to the commonly-used mT5, for zero-shot cross-lingual knowldge transfer in generation. We focus on encoder-decoder models, as they are well suited for generation, and consider mBART (pretrained and finetuned versions) and NLLB-200. We study two adaptation methods (full finetuning and parameter-efficient finetuning with adapters), pay attention to adaptation hyperparameters and compare models in a unified setting. We consider two tasks; summarization and questions answering (QA). Our main findings are as follows:
\begin{itemize}
    \item mBART performs best with adapters while for other models adapters and full finetuning perform similarly.
    \item mBART with adapters
    performs similarly to mT5 of the similar size. Qualitatively, due to the specifics of masking pretraining objective, mBART is better suited for tasks with long outputs while mT5 is for tasks with short outputs.
    \item 
    NLLB-200 is surprisingly competitive in summarization, reaching performance of mT5 and mBART  for high-resourse Latin-alphabet languages, but lags behind in QA.
    \item Hyperparameter tuning plays an important role: while most of the works report severe problems with generation in wrong language for mT5 with full finetuning, we find that simply reducing learning rate helps to alleviate this problem almost completely, without hurting performance.
\end{itemize}

\section{Related Work}
All works on zero-shot cross-lingual generation underline (and try to address) the severe problem of generating in a wrong language at the test time. This problem is also referred to under terms catastrophic forgetting (of languages not participating in finetuning, \citealp{vu-etal-2022-overcoming}), source language hallucination~\citep{mmt5}, or accidential translation problem~\citep{li-murray-2023-zero}.
\citet{vu-etal-2022-overcoming} 
propose to overcome generation in a wrong language
by
using parameter-efficient finetuning instantiated by prompt-tuning\footnote{Prompt tuning comprises prepending several learnable vectors ("prompt") to the list of embeddings of text input and freezing all other weights of model during finetuning.}, mixing-in the unsupervised target language task together with the supervised source language task, and factorizing learnable prompts into language and task components. They test these approaches on mT5.
\citet{mmt5} propose mmT5 (modular mT5), allocating a small amount of language-specific parameters in the model during pretraining and freezing them during task-specific finetuning. To alleviate 
generation in a wrong language,
they freeze some additional mmT5 parameters during finetuning, e.~g.~embedding layer and feed forward layers in Transformer decoder. \citet{li-murray-2023-zero} argue that learning language-invariant representations during finetuning is harmful for cross-lingual generation and propose finetuning on data from more than one source language
to avoid generation in a wrong language, with mT5 as a base model. 
ZMBART~\citep{zmbart} is the only work which considers other base model than mT5: they further pretrain mBART on an auxiliary unsupervised task on Hindi, Japanese and English. To avoid generation in a wrong language, they freeze embeddings and Transformer decoder, and mix-in data from auxiliary pretraining during finetuning. 
None of existing works compare different base models under a unified setting, which is a focus of this work.

Alternative approaches to zero-shot cross-lingual transfer include translate-train and translate-test paradigms. The former one implies translating task data to the target language and finetuning the model on this translated data, and the latter one assumes translating test input examples into the source language, generating outputs in the source language and translating them back into the target language. The drawbacks of these approaches include a high computational cost either at training or testing time, lack of high-quality translation models for low-resource languages, and potential inconsistencies between sentences in translation~\citep{vu-etal-2022-overcoming}.
Another related field is \textit{few-shot cross-lingual generation} which assumes access to a small amount of labeled examples in the target language~\cite{schmidt-etal-2022-dont,lauscher-etal-2020-zero, zhao-etal-2021-closer}.

\section{Methodology and experimental setup}

\paragraph{Models.}
We focus on encoder-decoder mPLMs as they are well suited and widely used for generation purposes, as opposed to encoder-only mPLMs such as mBERT or XLM-R. We leave the investigation of decoder-only mPLMs such as BLOOM~\citep{bloom} for future work. We consider mT5 and mBART as two most widely used mPLMs and NLLB-200 as a high-quality translation model:
\begin{itemize}
    \item mT5: pretrained using the masked language modeling objective where parts of the input sequence are masked and the missing fragments act as targets\footnote{In contrast to English-centric T5, mT5 did not include supervised tasks in pretraining.}. 
    mT5 is pretrained on the mC4 corpora, supports 101 languages, and does not use any language codes. Among released sizes from 300M to 13B we experiment with mT5-base (580M, most of the experiments) and mT5-Large (1.2B, additional experiment).
    \item mBART (pt): pretrained using the denoising objective where parts of the input sequence are masked and the entire original sequence acts as a target~\citep{mbartpt, mbarttr}. mBART is pretrained on Common Crawl~\cite{commoncrawl} corpora, supports 50 languages, has 680M parameters in total and uses language codes in both encoder and decoder sides. Both input sequence \verb|X| and target sequence \verb|Y| are prepended with the language code: \verb|[lang_code, X]| and \verb|[lang_code, Y]|, and at the inference time \verb|lang_code| is forced as a first generated token. Our hypothesis is that the use of the language code in the decoder can help to alleviate the problem of generation in a wrong language.
    \item mBART (tr): In addition to the \textit{pretrained} version, we also consider mBART finetuned for \textit{translation}~\citep{mbarttr}.
    \item NLLB-200: translation model supporting 200 languages, pretrained on sentence-level data mined from the web and automatically paired using multilingual embeddings. NLLB-200 uses the same language code scheme as mBART and is released in various sizes spanning from 600M to 54.5B, among them we consider 600M (distilled version, most of the experiments) and 1.3B (distilled version, additional experiment). Our hypothesis is that translation-based pretraining may provide good representations for cross-lingual transfer.
\end{itemize}

\paragraph{Adaptation methods.}
We consider full finetuning of the model and parameter-efficient finetuning with adapters~\citep{houlsby, bapna-firat-2019-simple}. Adapters are lightweight tuned
modules inserted after each fully-connected block of Transformer, when the rest of (pretrained) model weights are frozen.

\paragraph{Tasks and setting.}
We experiment with news summarization on the XL-Sum dataset~\citep{xlsum} (ROUGE metric~\cite{rouge}) and question answering on the XQuAD dataset~\citep{xquad} (F-measure metric). In the former task, the model needs to generate a 1--2 sentences summary based on a 1--2 news paragraphs, and in the latter task, the model needs to generate a short phrase answer based on a paragraph and question about it appended in the end of the paragraph.
For better metrics interpretability in cross-lingual QA, we compute metrics only over questions for which groundtruth answers do not contain numbers and are correctly identified to be written in the target language. We select a representative subset of languages for each task, covering Latin- and non-Latin scripts. We train models on English data for 20k steps with batch size of 4000 tokens on a single V100 or A100 GPU, and evaluate on validation sets of considered target languages each 2k steps. We grid search the learning rate (LR) for each task-model-adaptation method combination. We crop input sequences to the maximum length supported by models, which equals to 512 (mT5, NLLB-200) or 1024 tokens (mBART).
We report the task-specific metric and the percentage of outputs generated in the correct target language. More details on the experimental setting are given in Appendix~\ref{app:exp_details}.

\section{Experiments}

\begin{figure}[h!]
    \centering
    \begin{tabular}{c}
    Summarization (XL-Sum) \\
         \includegraphics[width=\linewidth]{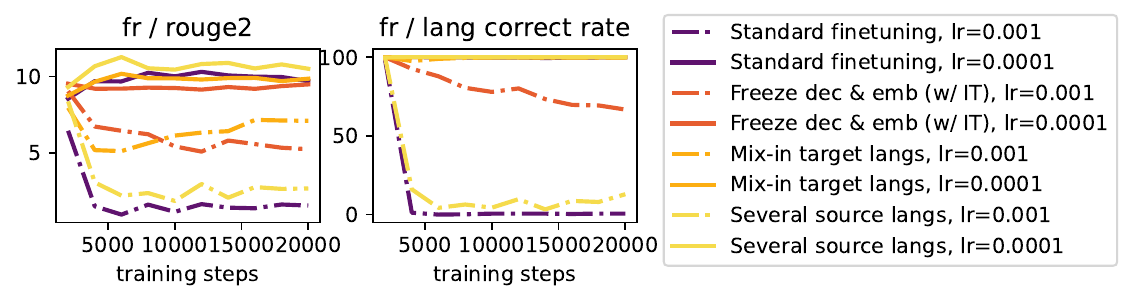}  \\
         Question answering (XQuAD) \\
          \includegraphics[width=\linewidth]{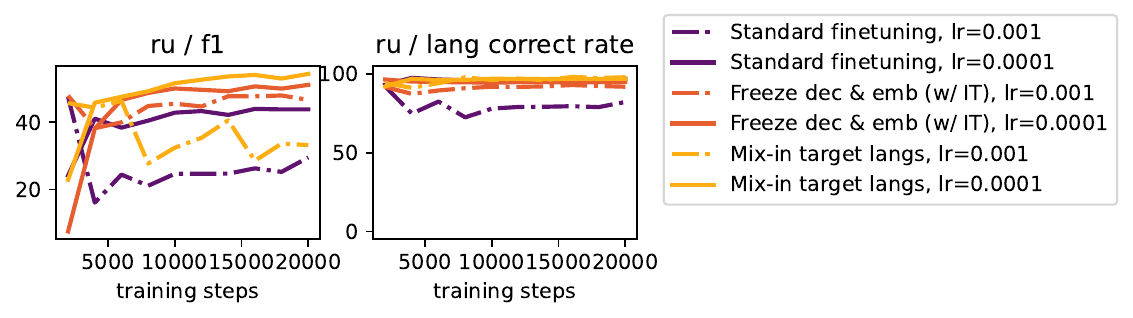}
        \end{tabular}
        \caption{Effect of learning rate for full finetuning of mBART (pt) and mT5. Russian as an example language, full results are presented in Appendix.}
        \label{fig:lr_ft}
\end{figure}

\paragraph{Effect of learning rate.} 
In Figure~\ref{fig:lr_ft} we show validation curves for full finetuning of mT5-base and mBART with various LRs, for Russian language as an example. Full set of plots demonstrating the effect of LR in each task-model-adaptation method-language combination is presented in Figures \ref{fig:mt5_xlsum}--\ref{fig:mbart_xquad} in Appendix. 
With too small or too large LR the model does not learn even the English task because of too short steps or divergence. For the range of LRs when the English task is learned well, we observe that larger LRs lead to the effect reported in other works, when the model overfits to the source English language and generates answers in English when applied in cross-lingual setting. 
However, \textit{with the reduced LR, this effect almost completely eliminates and the model mostly generates in the target language}. This trend is especially significant for full finetuning, where only small LRs enable correct language generation and slight increase in LR can substantially reduce metrics. Adapters in general enable higher LR magnitude without wrong language generation. 
In the following experiments, for each task-model-adaptation method combination, we choose (and report in Table~\ref{tab:lrs} in Appendix) the LR with the highest achieved task metric value averaged over non-English languages. This LR usually corresponds to the highest LR allowing generation in the correct language. We could not find information on the used LR in \citet{mmt5}, \citet{vu-etal-2022-overcoming}, to compare our chosen LRs with theirs. 

When reducing the LR for preserving generation in correct language, a reasonable question could be whether predictions of higher LR models are higher quality answers, but just in the wrong language, or simply hallucinations caused by data distribution shift. The premise for the former scenario is that on English data, performance with our chosen LR is usually slightly lower than with a larger LR. 
We find that actually the later scenario takes place, by comparing
performance of our chosen LR (best for non-English) and of the best LR for English with model predictions being translated into target languages using NLLB-3B\footnote{NLLB-3B handles well inputs containing code switching which are frequent in predictions we are translating, and simply copies inputs which are already in the target language.}, for last checkpoints of full models finetuning. According to Table~\ref{tab:translation}, translated predictions of higher LR model score lower than the (non-translated) predictions of lower LR model. This result further advocates for the importance of careful LR tuning for full finetuning in zero-shot cross-lingual generation. 

\begin{table}[]
\centering
\begin{small}
\begin{tabular}{p{0.3cm}l|cc|cc} 
\toprule
{} &{} &  \multicolumn{2}{c|}{\textbf{Best-En LR + Tr.}}  &   \multicolumn{2}{c}{\textbf{Best-non-En LR}}  \\ \midrule
{} &{} &  \textbf{LR} &    \textbf{Score} &    \textbf{LR} &  \textbf{Score} \\
\midrule
{} & mT5 & 1e-3  & 4.02  &  1e-4  & 7.7  \\
Sum & mBART & 1e-5  & 4.06  &  1e-6  & 5.34  \\
{} & NLLB-200 & 1e-4  & 2.86  &  1e-5  & 4.62  \\ 
\midrule
{} & mT5 & 1e-4  &  46.2 &  1e-4  & 58.6  \\
QA & mBART & 1e-5  & 41.1  &  1e-5  & 46.6  \\
{} & NLLB-200 & 1e-4  &  17.4 &  3e-5  &  18.2 \\
\bottomrule
\end{tabular}
\end{small}
    \caption{Comparison of best LR for non-English languages and best LR for English with model outputs being translated into target languages. Performance averaged over non-English languages, after 20k of full finetuning. Reported metric: Rouge-2 for summarization, F-measure for QA. mBART --- pretrained version.
    }
    \label{tab:translation}
\end{table}

\begin{figure*}[h!]
    \centering
         \includegraphics[width=\linewidth]{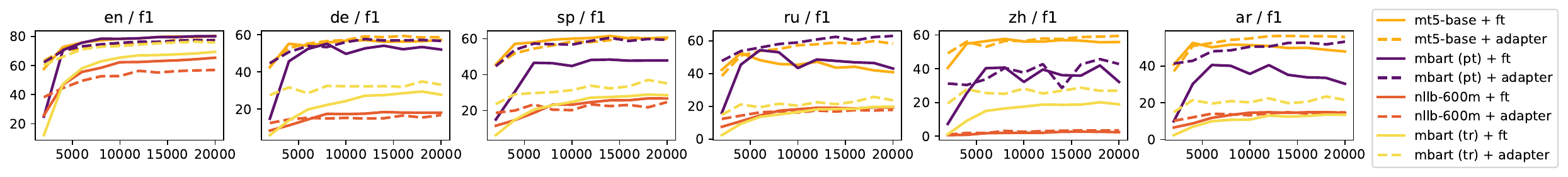} 
          \includegraphics[width=\linewidth]{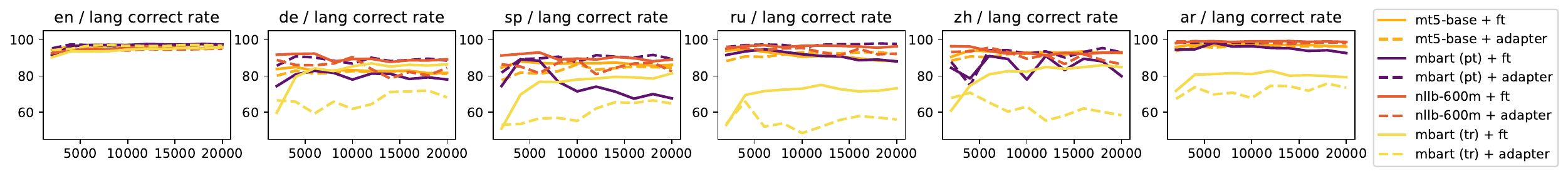} 
          \includegraphics[width=\linewidth]{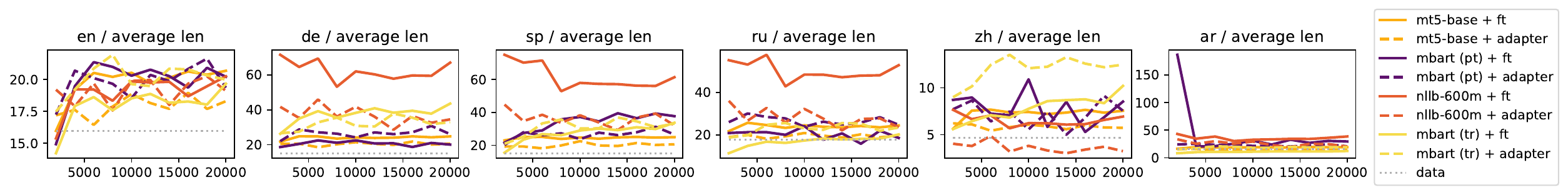} 
        \caption{Results for question answering (XQuAD dataset) in selected languages, x-axis represents training steps. Each plot averaged over 3 runs. Correct language rate is underestimated for Latin script languages (English, French, Spanish) because of language identification errors arising sometimes for very short answers, e.g. names. However, a low percent of wrong language predictions indeed can be found for these languages in manual inspection, same as for other languages.}
        \label{fig:qa}
\end{figure*}

\paragraph{Comparison of adaptation methods and models.}
In Figures~\ref{fig:main_sum} and \ref{fig:qa} we compare all four models we consider, trained with full finetuning or adapters on English data, using their best LRs. We find that \textit{for mT5, adapters and full finetuning usually perform similarly}, while \textit{for mBART, adapters substantially outperform full finetuning}. Equipped with adapters, \textit{mBART performs similarly to mT5 of the same size}, with one of these two models being best for each task-language pair.  
Translation-pretrained \textit{NLLB-200 performs well in summarization}, achieving performance of mT5 and mBART in Latin-language high-resource languages, French and Spanish, and performing on par with finetuned mBART in other languages\footnote{Expect Chinese, for which NLLB-200 generates a lot of empty predictions, as visible from the average length plots. NLLB-200 was noticed previously in the literature to have issues with processing Chinese.}. 
We inspected its predictions in Russian and French and they indeed form meaningful summaries. However, in QA, NLLB-200 performs poorly, often (but not always) generating non-relevant answers. Translation-finetuned version of mBART performs poorly in all tasks, generating a lot of wrong language predictions. We connect this to NLLB-200 being a much higher-performance translation model overall than mBART (tr). 

\begin{figure*}[h!]
    \centering
         \includegraphics[width=\linewidth]{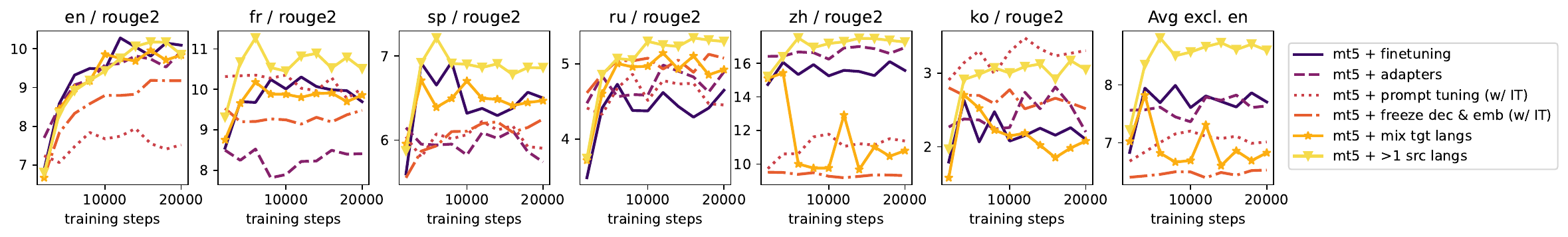} 
          \includegraphics[width=\linewidth]{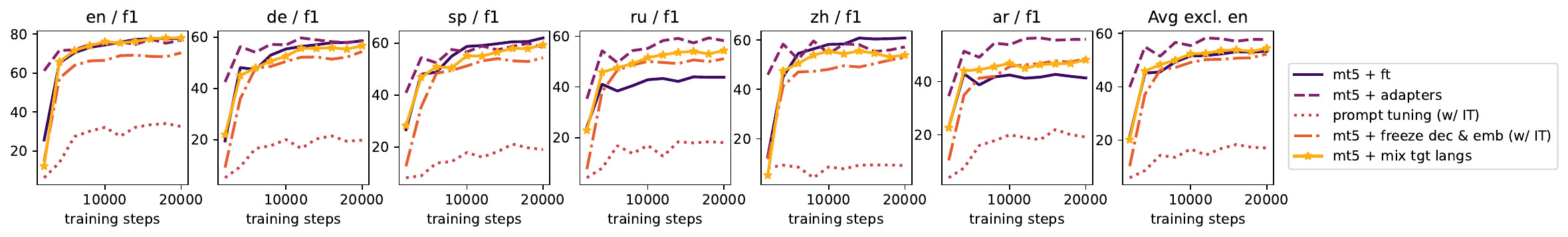} 
        \caption{Results for summarization (XL-Sum dataset) in selected languages, x-axis represents training steps. Each plot averaged over 3 runs. Not shown correct language plots for mBART (tr) are entirely below 50\% level.}
        \label{fig:main_sum}
\end{figure*}

\begin{figure*}[h!]
    \centering
         \includegraphics[width=\linewidth]{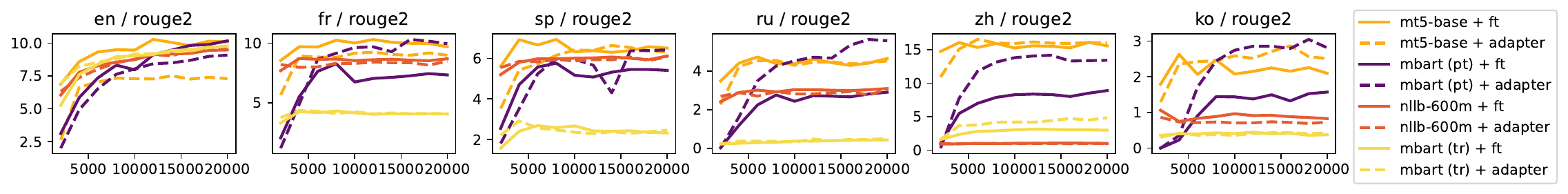} 
          \includegraphics[width=\linewidth]{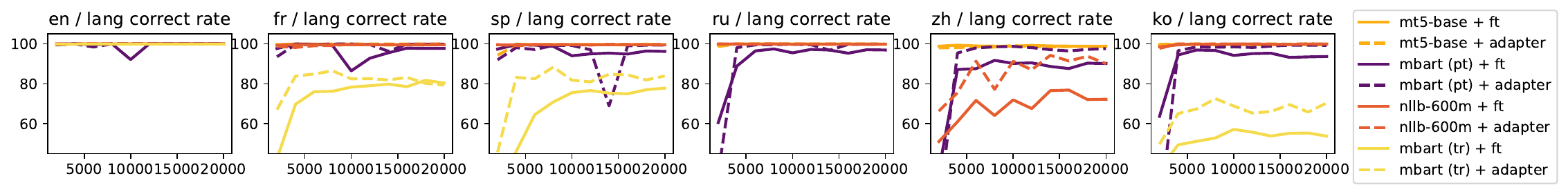} 
          \includegraphics[width=\linewidth]{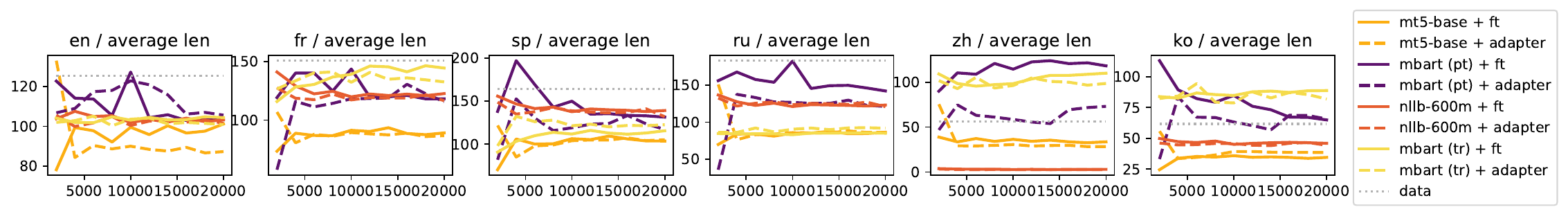} 
        \caption{Results for summarization (XL-Sum dataset) in selected languages, x-axis represents training steps. Each plot averaged over 3 runs. Not shown correct language plots for mBART (tr) are entirely below 50\% level.}
        \label{fig:main_sum}
\end{figure*}

\paragraph{Experiments with larger models.}
We test mT5-Large (1.3B params) and NLLB-200 (1.3B params) with the same learning rates as for smaller variants and find that their behavior is similar to smaller variants, 
e.g. similar curves for the correct language rate and similar trend for performance, 
but of course with higher absolute values of the task metric. These runs are demonstrated in Figures \ref{fig:mt5_xlsum}, \ref{fig:mt5_xquad}, \ref{fig:nllb_xlsum}, \ref{fig:nllb_xquad} in Appendix.

\paragraph{Inspection of predictions.} 
While inspecting predictions in Russian and French, we found that models achieving highest scores in both tasks generate fluent, meaningfull and reasonable predictions in a lot of cases, but sometimes have issues with truthfulness or hallucinations. Analysing effects of LR, we observe that increasing LR leads first to increase in code switching and then to wrong language generation, while \textit{reducing LR leads to producing rudiments of pretraining in generation}. For example, models sometimes generate extra tokens used in pretraining, such as \verb|<extra_id_{N}>| for mT5 or \verb|<sep>| for mBART, which we remove from predictions before computing metrics. \textit{In most cases this does not affect grammaticality of predictions}, but in rare cases leads to mT5 producing incomplete sentences, which may look unreasonable in summarization, e.g. generated summaries ``\texttt{<extra\_id\_0>, which can reproduce handwritten font on a pad of paper.}'' (translated from Russian) or ``\texttt{<extra\_id\_0> Guinea-Bissau President Alberto Dabo said.}'' (translated from French). The reason is that in mT5 pretraining tokens \verb|<extra_id_{N}>| were followed by fragments of input sentences.

In the same fashion, \textit{mBART average lengths are closer to groundtruth average lengths than mT5 in summarization, and the reverse effect takes place in QA}. This can be seen from average length plots in Figures~\ref{fig:main_sum} and \ref{fig:qa}. The reason is that in mT5 pretraining, the targets are only fragments masked in the input, which are shorter than targets in mBART pretraining represented by full sequences (they need to be reconstructed from the masked inputs).

\section{Conclusion}
In the same fashion, \textit{mBART average lengths are closer to groundtruth average lengths than mT5 in summarization, and the reverse effect takes place in QA}. This can be seen from average length plots in Figures~\ref{fig:main_sum} and \ref{fig:qa}. The reason is that in mT5 pretraining, the targets are only fragments masked in the input, which are shorter than targets in mBART pretraining represented by full sequences (they need to be reconstructed from the masked inputs).
In this work, we compared encoder-decoder multilingual pretrained language models in zero-shot cross-lingual generation, on summarizarion and question answering tasks. We found that the learning rate and the adaptation strategy (finetuning vs adapters) play an important role in the studied setting, and that with careful tuning, mBART performs similarly to the commonly used mT5 of the same size. Translation-pretrained NLLB-200 shows good performance in summarization but lags behind in question answering. The advantage of mT5 and NLLB-200 are that they are released in various model sizes, with larger models achieving higher absolute performance. We suggest that future works report more rigorously their experimental setup and details on hyperparameter search, and consider wider spectrum of models in experiments. An interesting future research direction would be to study the effect of hyperparameters in existing approaches for overcoming generation in a wrong language.

\section{Acknowledgments}
We gratefully appreciate Alexandre Bérard's help with mT5 implementation.

\bibliography{anthology,custom}

\appendix
\clearpage
\section{Experimental setup}
\label{app:exp_details}

\paragraph{Data.} 
We experiment with news summarization on the XL-Sum dataset~\citep{xlsum} and question answering on the XQuAD dataset~\citep{xquad}. The XL-Sum dataset was obtained by crawling BBC news in 44 languages, with corpus size per language varying from 1K (Scottish Gaelic) to 300K (English) article-summary pairs. Inputs are composed of 1--2 paragraphs and targets are usually 2--3 sentences. We evaluate on validation sets and crop validation sets larger than 1K samples, to 1K. The XQuAD dataset was obtained by translating SQuAD validation set~\cite{squad} into 11 languages, thus all language corpora are parallel. We use this dataset for evaluation and train on the training set of SQuAD (80K training instances). Each input is composed of a paragraph and a question about this paragraph appended in the end of the paragraph. Each output is an answer to a question, a short segment copied from the paragraph.

\paragraph{Preprocessing and postprocessing.} We tokenize data using each model's tokenizer. We crop model inputs and outputs to the maximum lengths supported by models, which equal to 1024 tokens for mBART and 512 tokens for mT5-base and NLLB-600M. Due to the design of pretraining, models may generate extra tokens such as \verb|<extra_id_{N}>| for or \verb|<sep>| for mBART. We remove such extra tokens from predictions before computing metrics.

\paragraph{Training.} We train models on English data for 20k steps with batch size of 4000 tokens on a single V100 or A100 GPU, and conduct validation on considered target languages each 2k steps. We use Adam optimizer with standard inverse square root LR schedule and warm up of 4k steps, and update model parameters after each mini-batch. Adapter dimension is set to 64. We experimented with a larger adapter dimension of 2048 for NLLB-200 in XQuAD which gave higher absolute scores but still much lower than other models.

\paragraph{Hyperparameter search.} For each task-model-adaptation method combination, we search the learning rate best for non English languages on average, looking at ROUGE-2 for summarization and F-measure for QA. We start with the set of three LRs: $\{10^{-k}, k=3, 4, 5\}$. If the optimal $k^* \ne 4$ then we extend search correspondingly to $k=2, 1$ or $k=6, 7$  until performance stops improving. For full finetuning, after we find optimal $k^*$ we also consider $3\cdot10^{-k^*}$. The motivation is that the optimal $k^*$ usually corresponds to the maximal $k$ that still allows generation in the correct language, and considering $3\cdot10^{-k^*}$ enables more accurate search for this maximum.

\paragraph{Evaluation.} For summarization, we report ROUGE-1/2/L~\cite{rouge}, and for QA, we report accuracy and F-measure. Due to space limitation, in the main text we report only ROUGE-2 and F-measure, other metrics are reported in Appendix. In QA, a lot of answers contain numbers or English words which could inflate metrics even if the model does not generate in the correct language. Moreover, the accuracy of language identification decreases on short answers, resulting in false indication of generation in wrong language. To avoid these issues, we compute metrics in QA only over questions for which groundtruth answers do not contain numbers and are correctly identified to be written in the target language ($\sim$50\% of 1190 questions satisfy this criteria).

To identify language, we use \verb|fasttext| library~\cite{fasttext1, fasttext2} and its \verb|lid.176.bin| model\footnote{\url{https://fasttext.cc/docs/en/language-identification.html}}.

\section{Full results with various learning rates}
\begin{figure*}
    \centering
     \includegraphics[width=\linewidth]{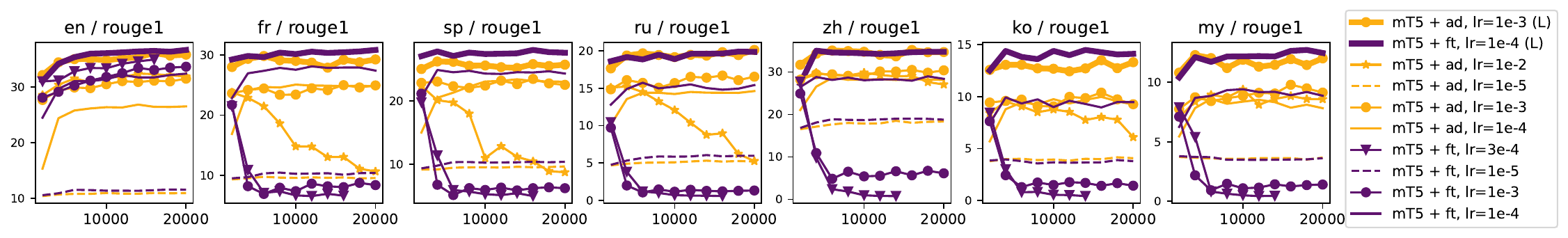} 
     \includegraphics[width=\linewidth]{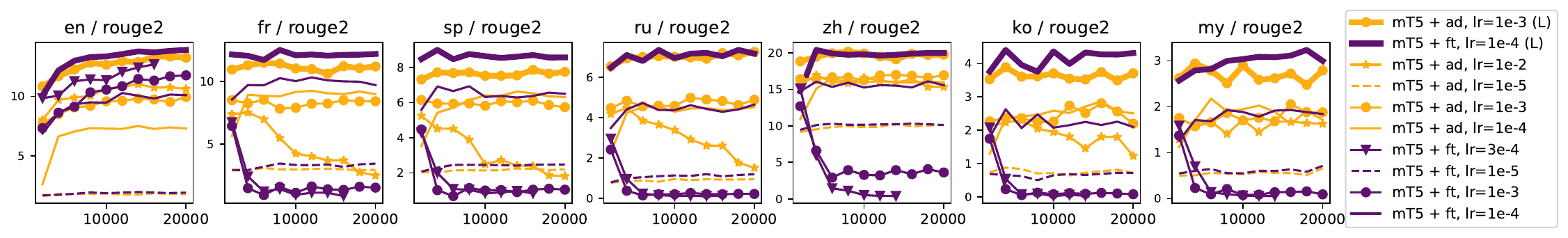} 
     \includegraphics[width=\linewidth]{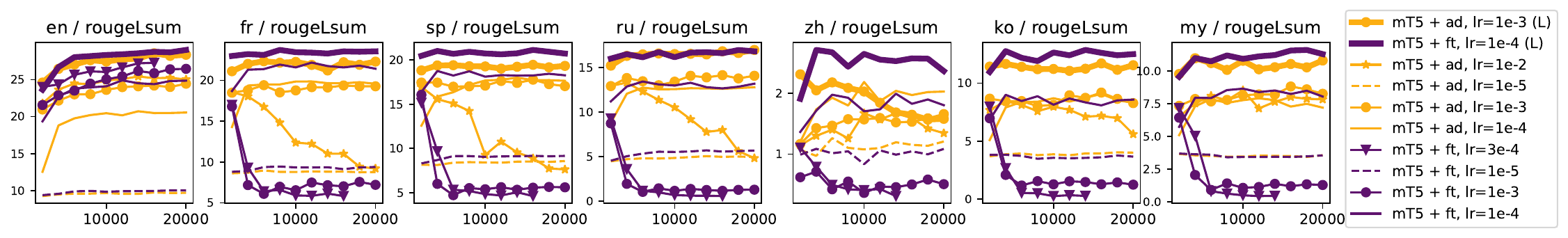} 
     \includegraphics[width=\linewidth]{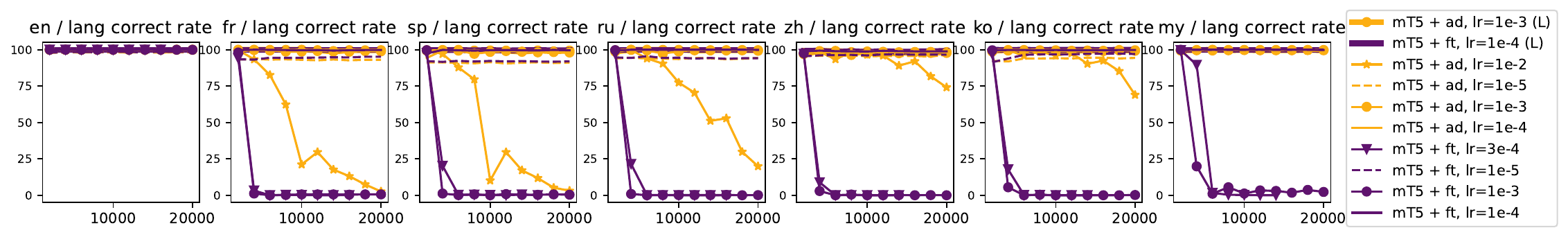} 
    \caption{Results for mT5 on XL-Sum with various learning rates. Lines marked with (L) represent mT5-Large, others represent mT5-base.}
    \label{fig:mt5_xlsum}
\end{figure*}

\begin{figure*}
    \centering
     \includegraphics[width=\linewidth]{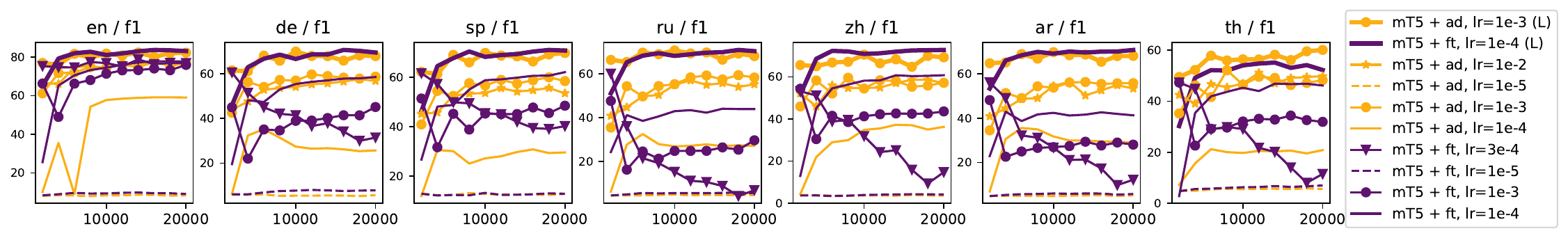} 
     \includegraphics[width=\linewidth]{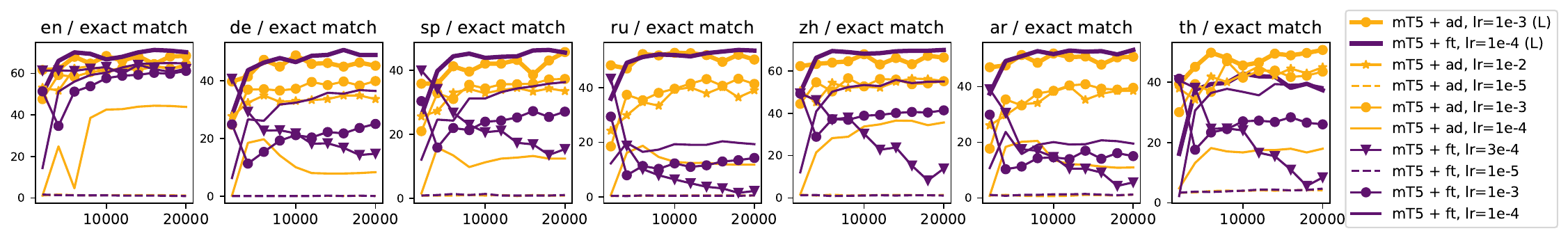} 
     \includegraphics[width=\linewidth]{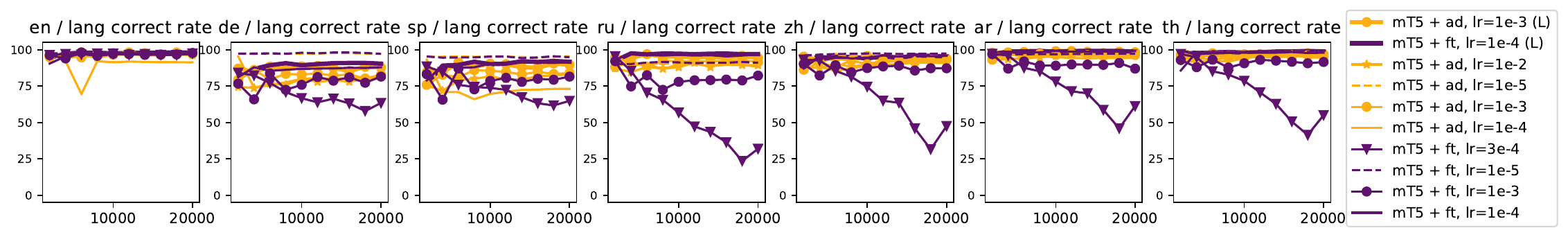} 
    \caption{Results for mT5 on XQuAD with various learning rates. Lines marked with (L) represent mT5-Large, others represent mT5-base.}
    \label{fig:mt5_xquad}
\end{figure*}

\begin{figure*}
    \centering
     \includegraphics[width=\linewidth]{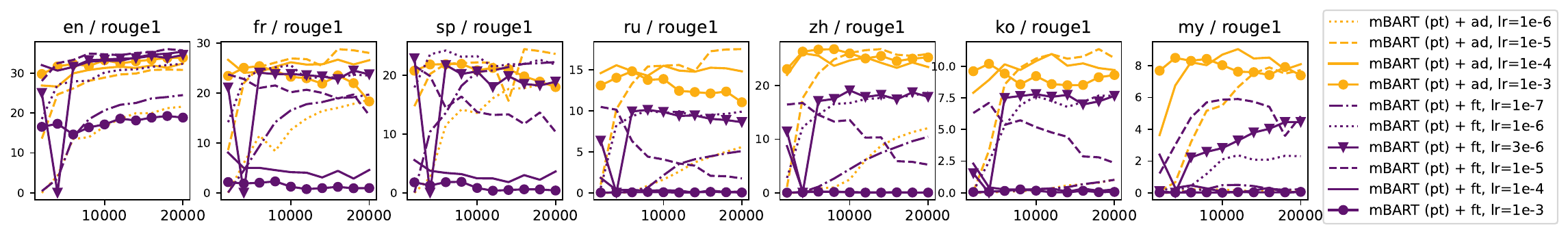} 
     \includegraphics[width=\linewidth]{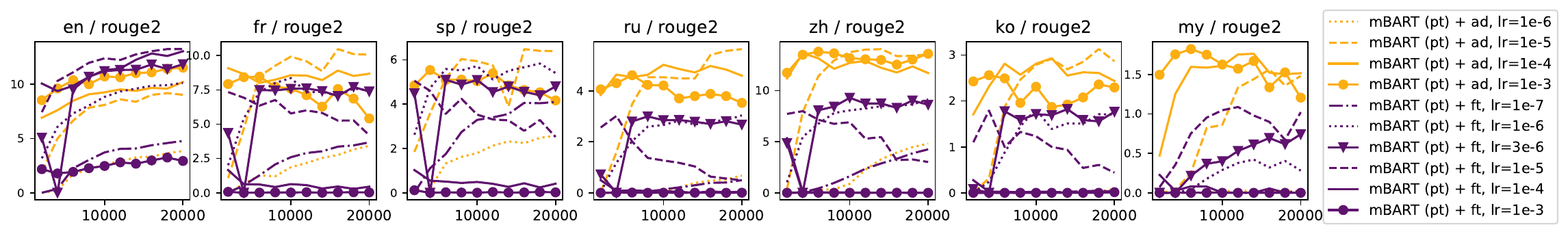} 
     \includegraphics[width=\linewidth]{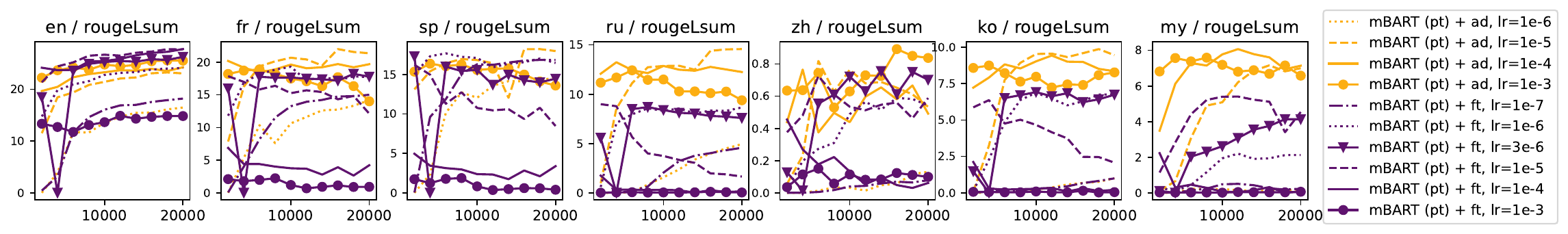} 
     \includegraphics[width=\linewidth]{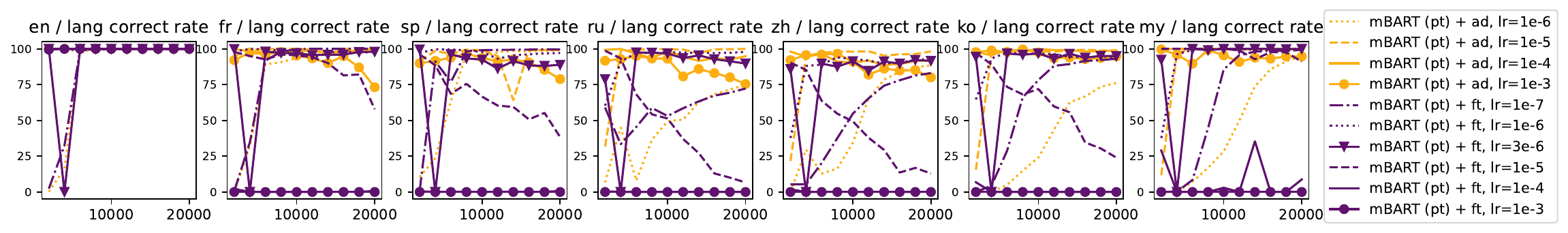} 
    \caption{Results for mBART-pt on XL-Sum with various learning rates.}
    \label{fig:mbart_pt_xlsum}
\end{figure*}

\begin{figure*}
    \centering
     \includegraphics[width=\linewidth]{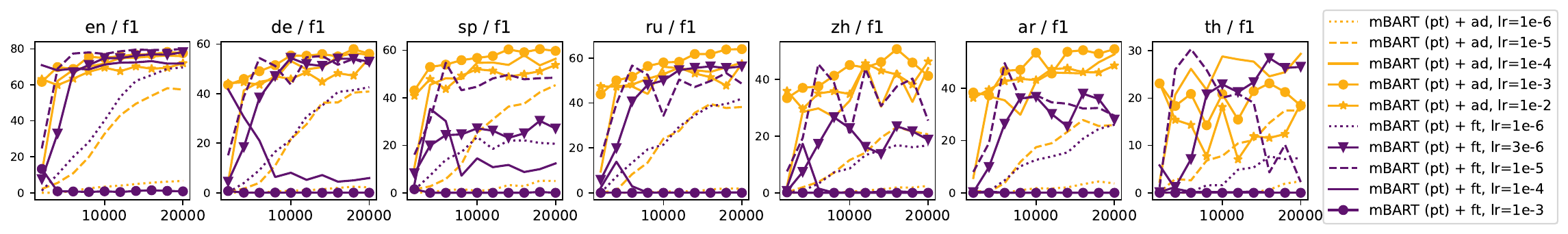} 
     \includegraphics[width=\linewidth]{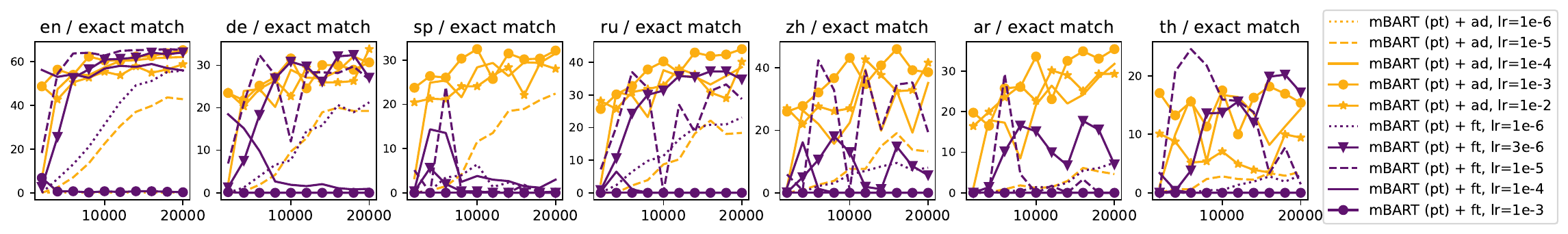} 
     \includegraphics[width=\linewidth]{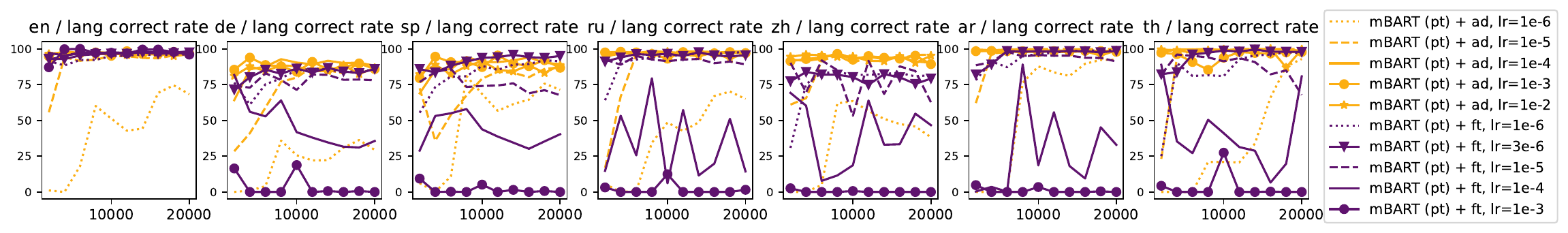} 
    \caption{Results for mBART-pt on XQuAD with various learning rates.}
    \label{fig:mbart_pt_xquad}
\end{figure*}

\begin{figure*}
    \centering
     \includegraphics[width=\linewidth]{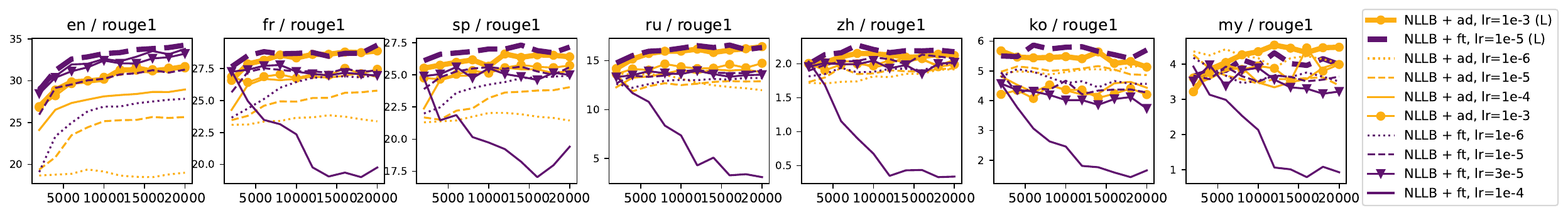} 
     \includegraphics[width=\linewidth]{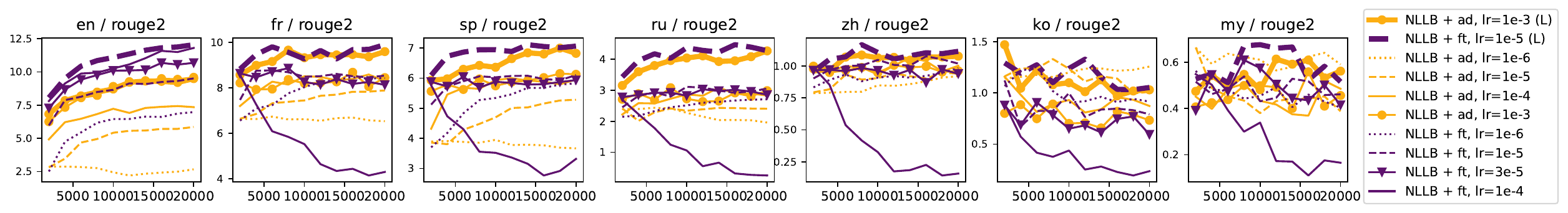} 
     \includegraphics[width=\linewidth]{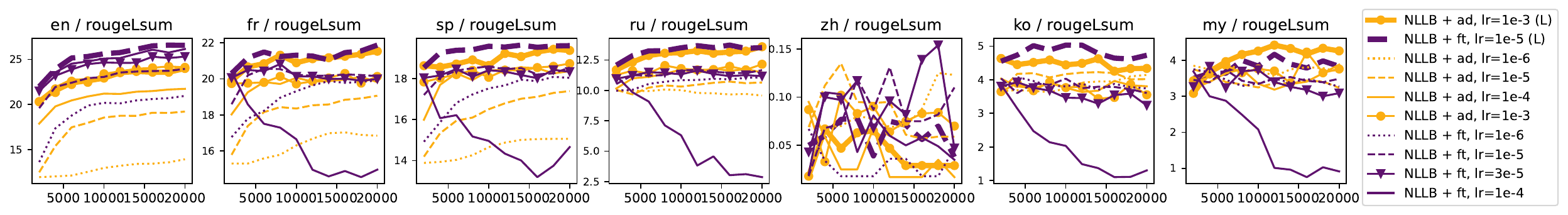} 
     \includegraphics[width=\linewidth]{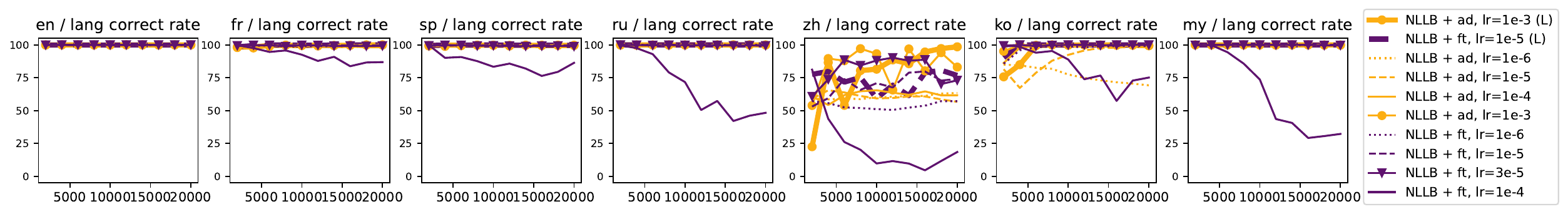} 
    \caption{Results for NLLB-200 on XL-Sum with various learning rates. Lines marked with (L) represent NLLB-1.3B, others represent NLLB-600M.}
    \label{fig:nllb_xlsum}
\end{figure*}

\begin{figure*}
    \centering
     \includegraphics[width=\linewidth]{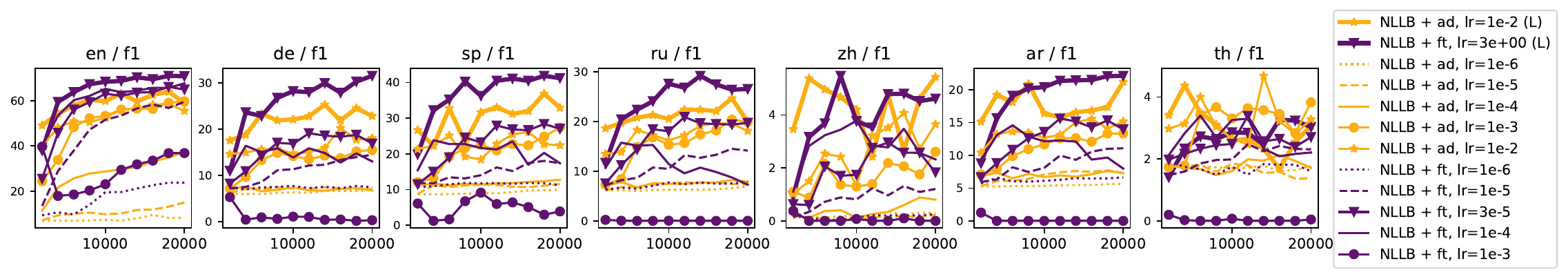} 
     \includegraphics[width=\linewidth]{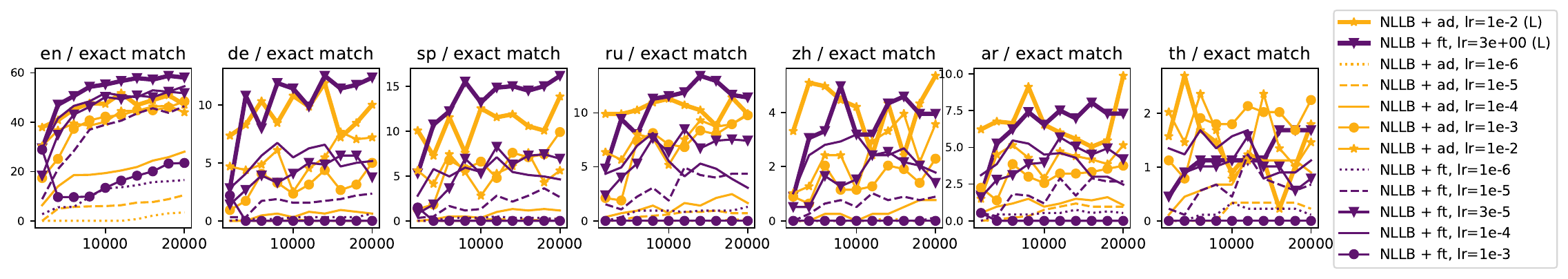} 
     \includegraphics[width=\linewidth]{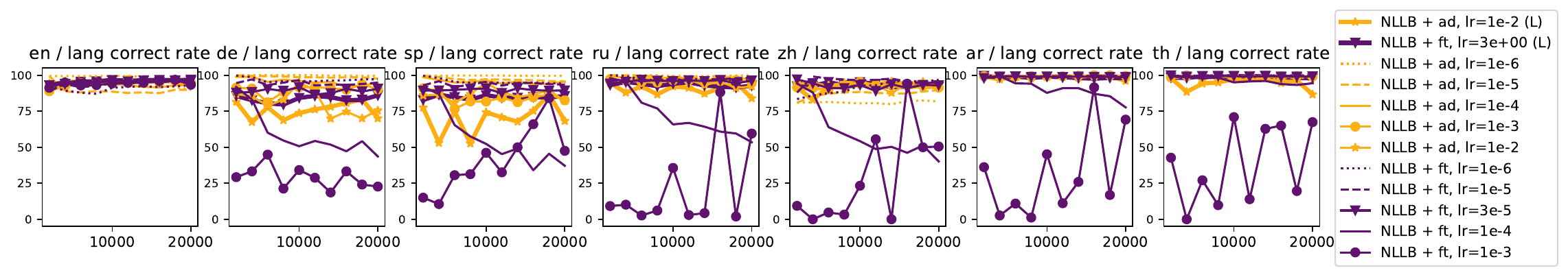} 
    \caption{Results for NLLB-200 on XQuAD with various learning rates. Lines marked with (L) represent NLLB-1.3B, others represent NLLB-600M.}
    \label{fig:nllb_xquad}
\end{figure*}

\begin{figure*}
    \centering
     \includegraphics[width=\linewidth]{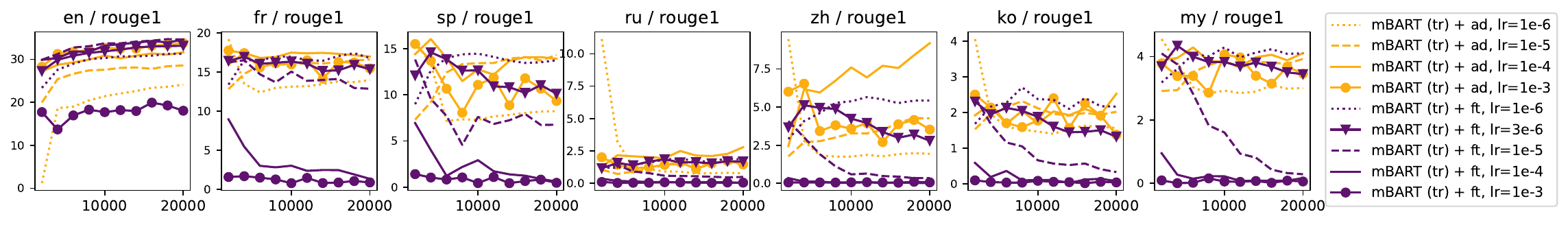} 
     \includegraphics[width=\linewidth]{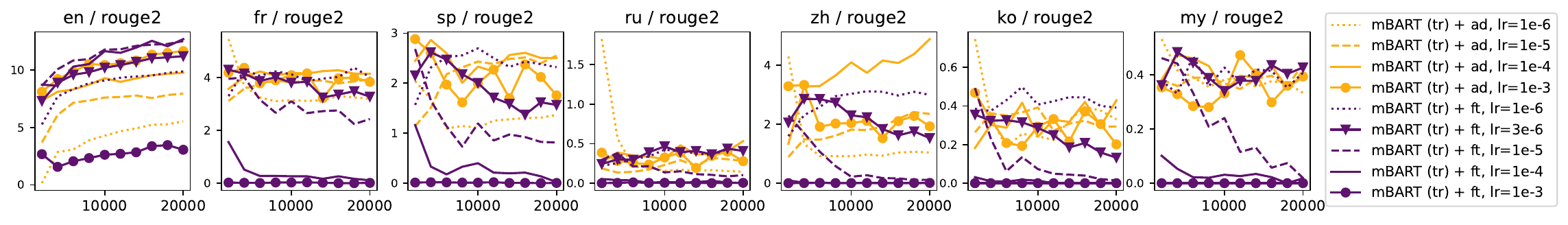} 
     \includegraphics[width=\linewidth]{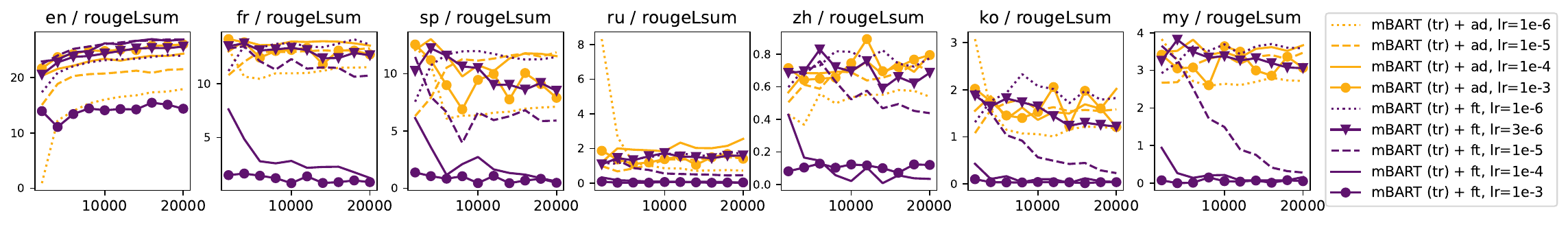} 
     \includegraphics[width=\linewidth]{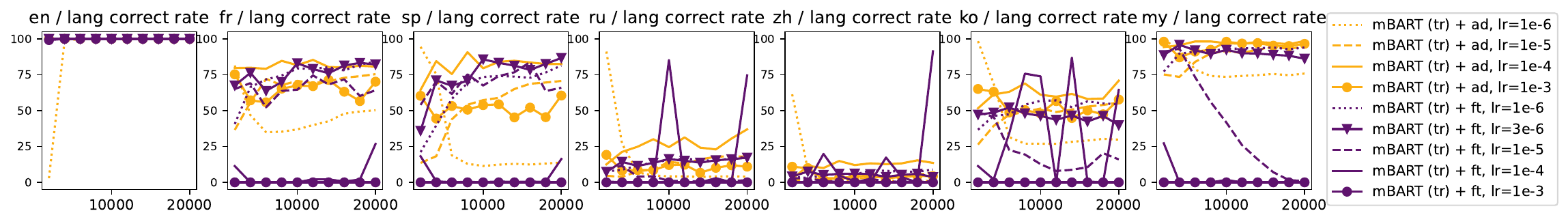} 
    \caption{Results for mBART-tr on XL-Sum with various learning rates.}
    \label{fig:mbart_xlsum}
\end{figure*}

\begin{figure*}
    \centering
     \includegraphics[width=\linewidth]{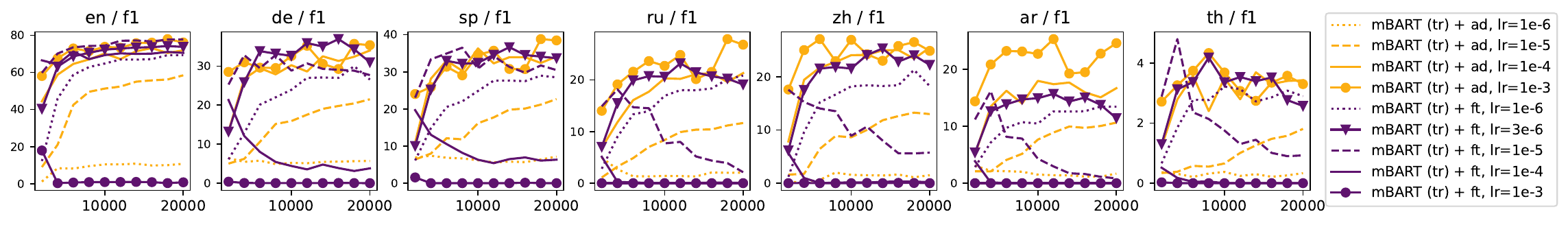} 
     \includegraphics[width=\linewidth]{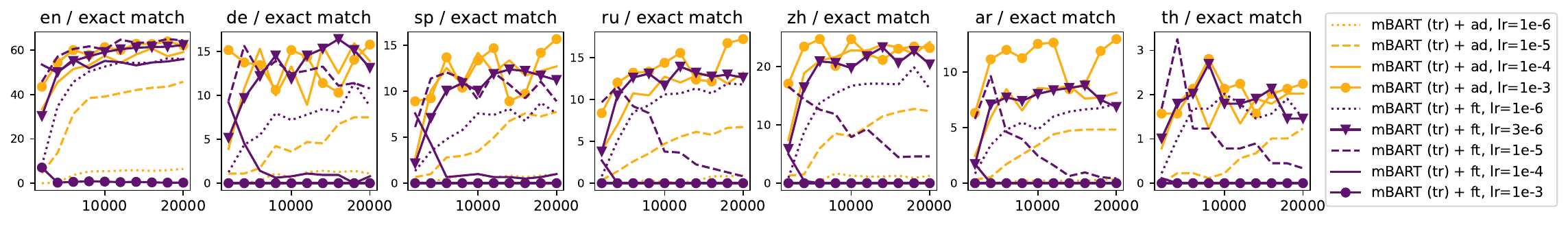} 
     \includegraphics[width=\linewidth]{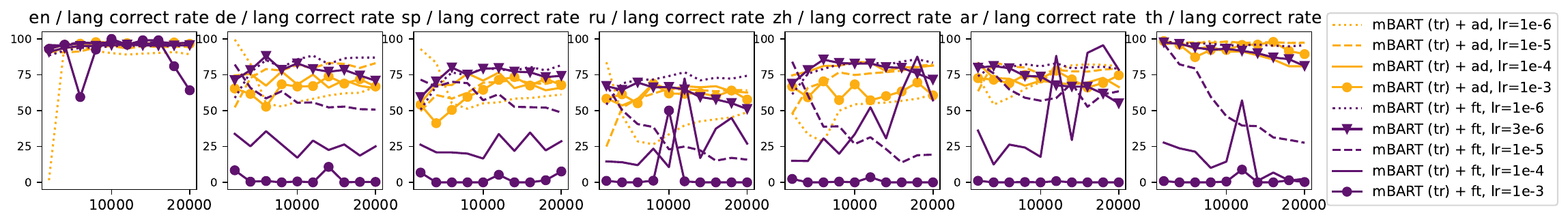} 
    \caption{Results for mBART-tr on XQuAD with various learning rates.}
    \label{fig:mbart_xquad}
\end{figure*}

Figures \ref{fig:mt5_xlsum}--\ref{fig:mbart_xquad} present full results for different models with two adaptation methods and various learning rates. In all figures, x-axis represents training steps.

\begin{table}[]
\centering
\begin{small}
\begin{tabular}{ll|cc} 
\toprule
\textbf{Task} & \textbf{model} &  \textbf{Full finetune} &    \textbf{Adapters} \\
\midrule
{} & mT5 &  1e-4 & 1e-4 \\
Sum & mBART-pt &  1e-6 & 1e-5 \\
 & mBART-tr & 1e-6  & 1e-4 \\
{} & NLLB-200 &  1e-5 & 1e-3 \\
\midrule
{} & mT5 &  1e-4 & 1e-3 \\
QA & mBART-pt & 1e-5  & 1e-3 \\
 & mBART-tr &  1e-3 & 3e-6 \\
{} & NLLB-200 & 3e-5  & 1e-2\\
\bottomrule
\end{tabular}
\end{small}
    \caption{Best learning rates for non-English languages.
    }
    \label{tab:lrs}
\end{table}

\end{document}